\documentclass{article}

\PassOptionsToPackage{numbers, compress}{natbib}

\usepackage[preprint]{neurips_2024}

\usepackage[utf8]{inputenc} 
\usepackage[T1]{fontenc}    
\usepackage{hyperref}       
\usepackage{url}            
\usepackage{booktabs}       
\usepackage{amsfonts}       
\usepackage{nicefrac}       
\usepackage{microtype}      
\usepackage{xcolor}         
\usepackage{animate}
\usepackage{subfig}
\usepackage{graphicx}
\usepackage{multirow}
\usepackage{amsmath}

\usepackage{algorithm}
\usepackage{algorithmicx}
\usepackage{algpseudocode}
\usepackage{wrapfig}
\usepackage{lipsum}
\usepackage{booktabs}
\usepackage{soul}

\title{Cinemo: Consistent and Controllable Image Animation with Motion Diffusion Models}

\author{
Xin Ma\textsuperscript{\rm 1,2$\dagger$}, 
Yaohui Wang\textsuperscript{\rm 2*}, 
Gengyu Jia\textsuperscript{\rm 3},
Xinyuan Chen\textsuperscript{\rm 2}, \\
\textbf{
Yuan-Fang Li\textsuperscript{\rm 1},
Cunjian Chen\textsuperscript{\rm 1*},
Yu Qiao\textsuperscript{\rm 2}}
 \\
\textsuperscript{1}Department of Data Science \& AI, Faculty of Information Technology, Monash University, \\
\textsuperscript{2}Shanghai Artificial Intelligence Laboratory,
\textsuperscript{3}Nanjing University of Posts and Telecommunications
}

\begin{document}
\renewcommand{\thefootnote}{}
\footnotetext{* corresponding authors. }
\footnotetext{$\dagger$ work done when Xin interned at Shanghai AI Laboratory}

\maketitle

\begin{abstract}

Diffusion models have achieved great progress in image animation due to powerful generative capabilities. However, maintaining spatio-temporal consistency with detailed information from the input static image over time (e.g., style, background, and object of the input static image) and ensuring smoothness in animated video narratives guided by textual prompts still remains challenging. In this paper, we introduce \textbf{Cinemo}, a novel image animation approach towards achieving better motion controllability, as well as stronger temporal consistency and smoothness. In general, we propose three effective strategies at the training and inference stages of Cinemo to accomplish our goal. At the training stage, Cinemo focuses on learning the distribution of motion residuals, rather than directly predicting subsequent via a motion diffusion model. Additionally, a structural similarity index-based strategy is proposed to enable Cinemo to have better controllability of motion intensity. At the inference stage, a noise refinement technique based on discrete cosine transformation is introduced to mitigate sudden motion changes. Such three strategies enable Cinemo to produce highly consistent, smooth, and motion-controllable results. Compared to previous methods, Cinemo offers simpler and more precise user controllability. Extensive experiments against several state-of-the-art methods, including both commercial tools and research approaches, across multiple metrics, demonstrate the effectiveness and superiority of our proposed approach. Project page: {\small \url{https://maxin-cn.github.io/cinemo_project}.}

\end{abstract}

\section{Introduction}
\label{introduction}

Image animation, also known as Image-to-Video generation (I2V), has persistently posed significant challenges within the realm of computer vision. It aims to convert an input static image into a video that exhibits natural dynamics while preserving the original detailed information of the input static image (e.g., architectural elements in the background or the artistic style of the input static image). Image animation has numerous real-world applications of interest such as photography, filmmaking, and augmented reality. 

Previous I2V approaches primarily focus on specific domains and benchmarks, for instance, human hair \cite{xiao2023automatic}, talking heads \cite{geng2018warp,wang2020imaginator,wang2022latent}, human bodies \cite{bertiche2023blowing,blattmann2021understanding,siarohin2021motion,wang2023leo}, etc., resulting in limited generalization capabilities in open domain scenes. Recently, with the success of large-scale diffusion models in image \cite{rombach2022high,chen2023pixart,podell2023sdxl}, 3D content \cite{lin2023magic3d,poole2022dreamfusion,li2023instant3d}, and video generation \cite{wang2023lavie,ma2024latte,blattmann2023align,luo2023videofusion,singer2022make,guo2024animatediff}, an increasing number of attempts have been made to extend such models into the realm of image animation \cite{zhang2023pia,dai2023animateanything,xing2023dynamicrafter,zhang2023i2vgen,ren2024consisti2v}, aiming to utilize the powerful content generation priors.

\begin{figure*}
\centering
\includegraphics[width=1.0\linewidth]{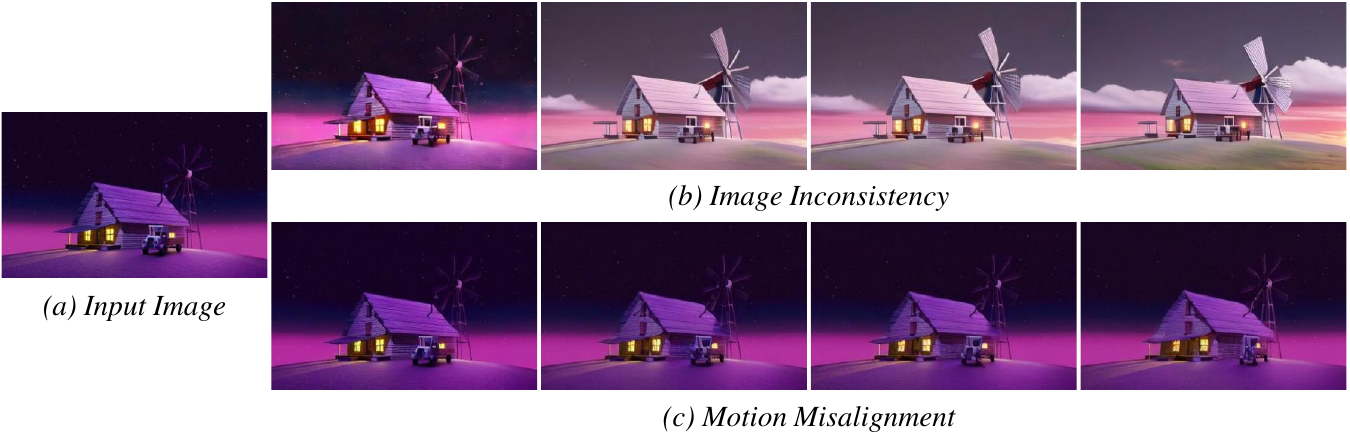}
\caption{\textbf{Explanations of image consistency and motion controllability.} Frames in (b) and (c) are image animation results obtained from PIA~\cite{zhang2023pia} and SEINE~\cite{chen2023seine}, respectively. We use \textit{``windmill turning''} as the text descriptions. (b) The frames show clear differences in color and texture. In (c), the entire house is moving, which does not match the information provided in the textual prompt.}
\label{fig_existing_methods_issues}
\end{figure*}

Initially, VideoComposer \cite{wang2024videocomposer} and VideoCrafter1 \cite{chen2023videocrafter1} utilize the CLIP \cite{radford2021learning} embedding of the input static image as an additional condition for Text-to-Video (T2V) model to achieve image animation. Subsequent works, such as AnimateLCM \cite{wang2024animatelcm}, DreamVideo \cite{wang2023dreamvideo}, and PIA~\cite{zhang2023pia}, observe that the CLIP image encoder tends to overlook details of the input static image, resulting in the model generating frames that do not capture detailed features of the original image. Consequently, these works adopt lightweight networks to extract the embedding of the input static image as the additional condition for the base T2V model. Other works, such as SEINE~\cite{chen2023seine} and VDT~\cite{lu2023vdt}, utilize a mask learning strategy to empower the model with the ability to animate images. However, as illustrated in Fig. \ref{fig_existing_methods_issues}, existing methods still face two major challenges: \emph{image consistency} and \emph{motion controllability}. Firstly, animated videos may fail to maintain consistency with the detailed information of the input static image over time. This manifests as significant discrepancies in dynamic visual elements such as shape, color, texture, etc., compared to the input static image, compromising the overall realism and coherence of the generated video (Fig. \ref{fig_existing_methods_issues} (b)). Secondly, existing methods may struggle to respond precisely to the motion patterns for a given 
textual prompt when generating the corresponding video, resulting in the generated motion sequences deviating from the context specified in the prompt (Fig. \ref{fig_existing_methods_issues} (c)).

To address the aforementioned limitations, we introduce \textbf{Cinemo}, a simple yet effective model that excels in both image consistency and motion controllability. Cinemo is designed based on a foundational T2V model \cite{wang2023lavie} to leverage its robust and powerful motion representation capability and incorporates the following novel designs. 

To endow our model with the ability to preserve the fine details of the input static image in the animated video, we design a novel learning strategy that focuses on learning the distribution of motion residuals rather than directly predicting subsequent frames, as commonly done by existing methods. At each diffusion time step, the appearance information of the input static image is added to the noised motion residual and then concatenated with the appearance information before being input into the model. The proposed model leverages the input static image to guide motion residual prediction effectively and directs the attention of the model toward generating the motion residual aligned with the provided prompt. Previous research typically relies on the \textit{frame per second} (FPS) \cite{chen2023videocrafter1} count or optical flow \cite{zhai2021optical} as a global measure to adjust the motion intensity in generated videos. However, FPS does not accurately reflect motion intensity; for instance, a video with high FPS may depict a nearly static scene. Although optical flow is able to assess motion intensity precisely, it encounters high computational costs and time consumption. To address this issue, we introduce a simple yet effective strategy based on the Structural Similarity Index (SSIM) \cite{wang2004image} to achieve fine-grained control over motion intensity. During the inference phase, we propose DCTInit which utilizes the low-frequency coefficients of the Discrete Cosine Transform of the input static image as layout guidance to refine the initial inference noise. DCTInit is able to address the noise discrepancy issue between training and inference phases as mentioned in previous works \cite{si2024freeu, wu2023freeinit, ning2023input} and mitigate sudden motion changes as shown in Fig. \ref{fig_effective_dctinit}. 

By integrating the three novel strategies, our approach can produce highly consistent videos from the input static image that closely aligns with the given prompt, and can easily be extended to other applications such as video editing and motion transfer. We conduct comprehensive quantitative and qualitative experiments and demonstrate that our model achieves state-of-the-art performance. We summarize our contributions as follows:

\begin{itemize}
\item We propose Cinemo, a diffusion-based image animation model with a focus on learning the distribution of motion residuals rather than directly predicting the next frames, avoiding video content distortion.

\item Towards mitigating sudden or undesired motion changes in animated videos, we introduce DCTInit, a strategy for refining initial inference noise during the inference phase, which utilizes the low-frequency Discrete Cosine Transform coefficients of the input static image. Additionally, an effective SSIM-based strategy for fine-grained control of video motion intensity is proposed to enhance motion intensity controllability in image animation.

\item  We conduct extensive quantitative and qualitative experiments to evaluate our model. The results demonstrate that our approach outperforms other methods in terms of image consistency and motion controllability.
\end{itemize}

\section{Related Work}
\label{related_work}
\textbf{Text-to-Video Generation} aims to produce high-quality videos by utilizing text descriptions as conditional inputs. In recent years, diffusion models \cite{ho2020denoising,song2021denoising,song2021score} and autoregressive models have made significant achievements in text-to-image (T2I) generation. Existing T2I methods are capable of generating realistic images closely aligned with textual prompts \cite{ramesh2021zero,saharia2022photorealistic,yu2022scaling,ramesh2022hierarchical}. Current text-to-video (T2V) generation approaches primarily involve augmenting T2I methods with additional temporal modules, such as temporal convolutions or temporal attentions, to establish temporal relationships between video frames \cite{wang2023lavie,guo2024animatediff,ma2024latte,wu2023tune,blattmann2023align,ge2023preserve,zhou2022magicvideo,he2022latent,villegas2022phenaki}. Due to the scarcity of available high-resolution clean video data, most of these methods rely on joint image-video training \cite{ho2022video} and typically build their models on pre-trained image models (such as Stable Diffusion \cite{rombach2022high}, DALL$\cdot$E2 \cite{ramesh2022hierarchical}, etc.). 
From the perspective of model architecture, current T2V techniques primarily focus on two designs: one is a cascaded structure \cite{ho2022imagen,ge2023preserve,singer2022make} inspired by \cite{ho2022cascaded}. The other is based on latent diffusion models \cite{blattmann2023align,zhang2023i2vgen,zhou2022magicvideo} extending the success of \cite{rombach2022high} to the video domain. From the perspective of the model purpose, these approaches can be mainly categorized into two major classes: firstly, most methods aim at learning general motion representations, which typically rely on large and diverse datasets \cite{wang2023lavie,ma2024latte,blattmann2023align,blattmann2023stable}; secondly, another well-recognized branch of T2V methods tackle the realm of personalized video generation, where they focus on fine-tuning the pre-trained T2I models on narrow datasets customized for specific applications or domains \cite{guo2024animatediff,wu2023tune}.

\textbf{Image Animation} aims to maintain the identity of the static input image while crafting a coherent video and has garnered attention and effort in the research field for decades. Early physics-based simulation methods focused on mimicking the motions of certain objects, but their lack of generalization stemmed from separately modeling each object category \cite{dorkenwald2021stochastic,prashnani2017phase,siarohin2021motion}. Subsequent GAN-based approaches overcome manual segmentation, enabling the synthesis of more natural movements \cite{wang2022latent,wu2021f3a,chen2020animegan,pumarola2018ganimation}. Currently, mainstream methods mainly rely on foundational T2I or T2V pre-trained models, using RGB images as additional conditions to generate video frames from input images. Some mask-based methods, such as SEINE \cite{chen2023seine} and VDT \cite{lu2023vdt}, employ random masking of input frames during training, which can predict future video frames from a single image. Plug-to-play methods like I2V-adapter \cite{guo2023i2v} and PIA \cite{zhang2023pia} utilize pre-trained LoRA~\cite{hu2022lora} weights to animate input images. While these methods are good at specific domains, they cannot guarantee continuity with given images. Additionally, there are animation methods focused on human bodies, such as AnimateAnyone \cite{hu2023animate} and MagicAnimate \cite{xu2024magicanimate}, which use additional motion sequences (i.e., poses) to drive the movement of a human body image. Although these methods achieve good video quality, they are limited to animating human body images and cannot animate other types of images. Furthermore, some commercial large-scale models like \href{ https://runwayml.com/ai-magic-tools/gen-2/}{Gen-2}, \href{https://www.genmo.ai/}{Genmo}, and \href{https://www.pika.art/}{Pika Labs} can produce realistic results in video frame quality but often fail to respond accurately to given textual prompts.

\begin{figure}
    \centering
    \includegraphics[width=1.0\linewidth]{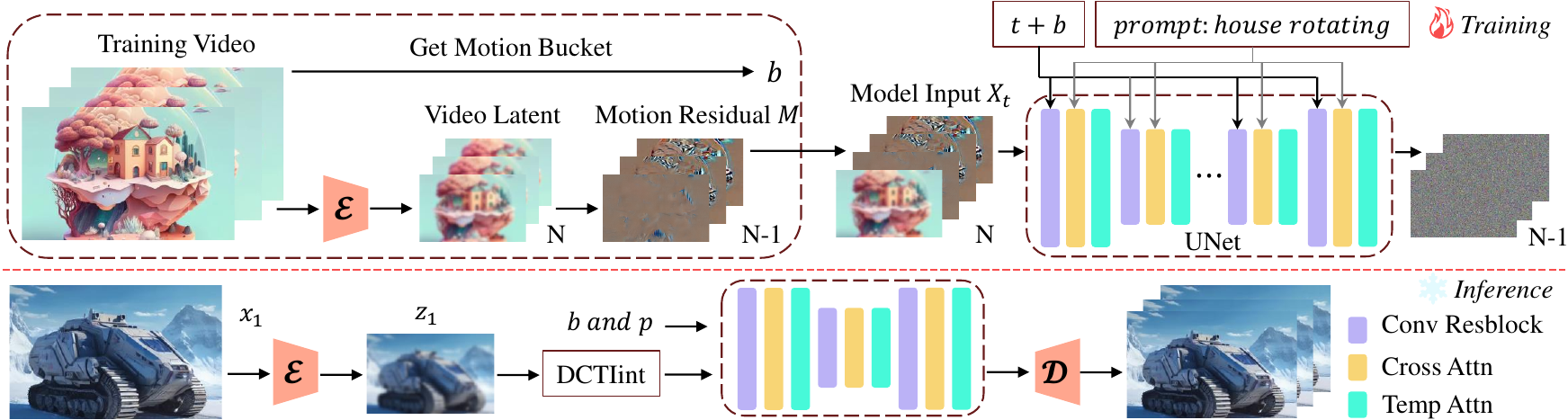}
    \caption{\textbf{Model pipeline overview.} During training, instead of predicting the subsequent frames directly, our model learns the distribution of motion residuals, while providing effective motion intensity control. The details of the training procedure can be seen in Algorithm. \ref{alg:training}. During inference, we use Discrete Cosine Transformation to extract low-frequency components to refine the inference noise, which can stabilize the generation process of image animation.}
    \label{fig_pipeline}
\end{figure}

\section{The Cinemo Image Animation Model}
\label{methodology}

We commence with a brief introduction of latent diffusion models, video latent diffusion models, and image animation formulation in Sec. \ref{method:preliminary}. Following that, we present the motion residual learning and motion intensity control in Sec. \ref{method_image_animation_formulation}. Finally, the DCT-based inference noise refinement strategy is discussed in Sec. \ref{method_dct_based_noise_refinement}. The overall pipeline of our model is shown in Fig. \ref{fig_pipeline}.

\subsection{Preliminary and problem formulation}
\label{method:preliminary}
\textbf{Latent diffusion models} (LDMs) are efficient diffusion models that operate the diffusion process in the latent space of the pre-trained variational autoencoder (VAE) rather than the pixel space~\cite{song2021denoising,rombach2022high,ho2020denoising,kingma2013auto,kingma2019introduction}. An encoder $\mathcal{E}$ from the pre-trained VAE is firstly used in LDMs to project the input data sample $x \in p_{data}$ into a lower-dimensional latent code $z = \mathcal{E}(x)$. Then, the data distribution is learned through two key processes: diffusion and denoising. The diffusion process gradually adds Gaussian noise into the latent code $z$ and the perturbed sample $z_{t} = {\sqrt{\overline{\alpha}_{t}}}z + \sqrt{1-{\overline{\alpha}_{t}}}\epsilon$, where $\epsilon\sim \mathcal{N}(0,1)$, following a $T$-stage Markov chain, is obtained. Here, $\overline{\alpha}_{t}$ and $t$ represent the pre-defined noise scheduler and the diffusion timestep, respectively. The denoising process learns to inverse the diffusion process by predicting a less noisy sample $z_{t-1}$: $p_\theta(z_{t-1}|z_t)=\mathcal{N}(\mu_\theta(z_t),{\Sigma_\theta}(z_t))$ and make the variational lower bound of log-likelihood reduce to $\mathcal{L_\theta}=-\log{p(z_0|z_1)}+\sum_tD_{KL}((q(z_{t-1}|z_t,z_0)||p_\theta(z_{t-1}|z_t))$. In this context, $\mu_\theta$ means a denoising model $\epsilon_{\theta}$ and is trained with the \emph{simple} objective,
\begin{align}
\label{equ_l_simple}
\mathcal{L}_{simple} = \mathbb{E}_{\mathbf{z}\sim p(z),\ \epsilon \sim \mathcal{N} (0,1),\ t}\left [ \left \| \epsilon - \epsilon_{\theta}(\mathbf{z}_t, t)\right \|^{2}_{2}\right].
\end{align}

\textbf{Video latent diffusion models} (VLDMs) expend LDMs into a video counterpart by introducing the temporal motion module to build the temporal relationship between frames \cite{wang2023lavie,ma2024latte,chen2023seine,wang2024videocomposer}. Our model is based on LaVie~\cite{wang2023lavie}, utilizing its pre-trained weights for initialization, and optimizes all parameters during training. Given a video clip $\mathbf{V} \in \mathbb{R}^{N \times C \times H \times W}$, where $N, C, H, W$ represent the total frames, channel numbers, height, and width respectively, we first project it into a latent space frame-by-frame to obtain latent code. After that, the diffusion and denoising processes are conducted in the latent space. Finally, the animated videos are generated by the decoder.

\textbf{Image animation formulation.} Given an input static image $x_1 \in \mathbb{R}^{C \times H \times W} $ and a textual prompt $p$, image animation aims to generate an $N$-frame video clip $\mathbf{V} = \{x_1, x_2, x_3, ..., x_N\}$, where $x_1$ is the first frame of the video and $x_i\in\mathbb{R}^{C \times H \times W}$. The appearance of the subsequent frames $\{x_2, x_3, ..., x_N\}$ should be closely aligned with $x_1$, while the content and motion of the video adhere to the textual description provided in $p$. We decompose this problem into three sub-problems: learning motion residuals, controlling motion intensity of generated videos (Sec.~\ref{method_image_animation_formulation}), and mitigating sudden motion changes of generated videos (Sec.~\ref{method_dct_based_noise_refinement}).

\subsection{Motion residual learning and motion intensity control} 
\label{method_image_animation_formulation}

\textbf{Motion residuals learning.} We propose a new learning strategy that focuses on learning motion residuals instead of directly predicting the subsequent frames as previous methods do, such as PIA~\cite{zhang2023pia}, ConsistI2V~\cite{ren2024consisti2v}, I2V-Adapter~\cite{guo2023i2v}, and DynamiCrafter~\cite{xing2023dynamicrafter}. This approach effectively alleviates the challenges faced by previous methods, including poor video frame quality, inconsistencies in the fine details of input images, and the issue of generating static videos (as shown in Figure \ref{fig:qualitative_comparison},). Consequently, it creates dynamic videos with higher frame quality. Specifically, we sample an $N$-frame video clip $\mathbf{V} = \{x_1, x_2, x_3, \ldots, x_N\}$, from the training dataset and set the first frame $x_1$ as the image to be animated. As described in Sec.\ \ref{method:preliminary}, we first use the encoder of the pre-trained VAE to compress the video clip $\mathbf{V}$ into a low-dimensional latent space to obtain the latent code $\mathbf{Z} = \{z_1, z_2, z_3, ..., z_N\}$, where $z_i\in\mathbb{R}^{c\times h\times w}$. Here, $c$, $h$, and $w$ indicate the channel, height, and width of the frame in latent space.


From $\mathbf{Z}$, we compute the \emph{motion residuals} $\mathbf{M}=\{z_2-z_1,z_3-z_1,\ldots,z_N-z_1\}$, by subtracting the first frame from all subsequent frames. To guide the model to predict motion residuals, at each diffusion timestep $t$, we first add the noised motion residuals $\mathbf{M}_t$ to the features of the input image $z_1$ to obtain $\mathbf{M}_t^{'}$. Then, we concatenate $z_1$ and $\mathbf{M}_t^{'}$ to form $N$ frames, which are used as the input $\mathbf{X}_t$ to the model. We select the last $N - 1$ frames output by the model as the denoised $\mathbf{M}_{t-1}$. The detailed procedure of our model is summarized in Algorithm. \ref{alg:training}.

\begin{wrapfigure}{r}{0.5\textwidth} 
\begin{minipage}{\linewidth}
\begin{algorithm}[H]
\caption{The training procedure. We assume that batch size is set to 1.}
\label{alg:training}
\begin{algorithmic}[1]
\Repeat
    \State Sample video $\mathbf{V} =\{x_1,\ldots,x_N\}$ from training set, where $x_i\in \mathbb{R}^{C \times H \times W}$ is a frame
    \State Compute motion bucket $b$ from $\mathbf{V}$ (Eq.\ \ref{eq:b})
    \State Compute $\mathbf{Z} = \{z_1, \ldots, z_n\}$, where $z_i=\mathcal{E}(x_i)\in\mathbb{R}^{c \times h \times w}$ for each $x_i\in\mathbf{V}$
    \State Compute $\mathbf{M} = \{z_2-z_1, z_3-z_1, \ldots, z_N-z_1  \}$
    \State Sample $t \sim \text{Uniform}(1,\ldots,T)$
    \State Sample $\epsilon \sim \mathcal{N}(0, I)$
    \State Get noised $\mathbf{M}_t$ via the diffusion process
    \State Get input $\mathbf{X_t}$ = \textit{torch.cat}([$z_1$, $\mathbf{M}_t$+$z_1$])
    \State Take gradient descent step on Eq.\ \ref{equ_l_final}
\Until{Converged}
\end{algorithmic}
\end{algorithm}
\end{minipage}
\end{wrapfigure} 

\textbf{Motion intensity controllability.} We propose a simple and effective strategy that uses the average structural similarity index (SSIM) $s$ between frames as a means to fine-grained control motion intensity:
\begin{equation}
    s(\mathbf{V}) = \frac{1}{N-1}\sum_{i=2}^N SSIM(x_i, x_{i-1}).\label{eq:b}
\end{equation}
Here, the motion intensity $s$ measures the differences between frames in the pixel space. We find that the motion intensities calculated from video clips sampled with a fixed frame interval exhibit a significant long-tail distribution. To alleviate this skewness, we randomly select a frame interval between 3 and 10 to sample video clips. After obtaining the motion intensity $s$, we uniformly project $s$ into the motion intensity bucket $b$ (ranging from 0 to 19). Similar to the timestep $t$, we project the motion intensity bucket $b$ onto the positional embedding, then add it to the timestep embedding, and finally incorporate it into each frame in the residual block to ensure that the motion intensity is applied uniformly across all frames.


Finally, combining \textit{motion residuals learning}, \textit{motion intensity controllability}, and Eq.\ \ref{equ_l_simple}, the final learning objective can be formulated as:
\begin{equation}
\label{equ_l_final}
\mathcal{L}_{final} = \mathbb{E}_{\mathbf{z}\sim p(z),\ \epsilon \sim \mathcal{N} (0,1),\ t}\left [ \left \| \epsilon - \epsilon_{\theta}(\mathbf{X}_t, p, b, t)\right \|^{2}_{2}\right].
\end{equation}

\subsection{DCT-based noise refinement}
\label{method_dct_based_noise_refinement}
During the training and inference phases, the noise input for the model $\epsilon_\theta$ is different. Specifically, as detailed in Sec. \ref{method:preliminary}, during the training phase, the model receives $z_t$ as input, which is obtained by sampling from the dataset and through $z_{t} = {\sqrt{\overline{\alpha}_{t}}}z + \sqrt{1-{\overline{\alpha}_{t}}}\epsilon$. In contrast, during the inference phase, the model obtains $z_t$ based on previous predictions. This discrepancy, also known as exposure bias or information leak~\cite{lin2024common,si2024freeu}, can lead to the accumulation of inference errors. 
In video generation, this difference primarily stems from the low-frequency components as a result of frequency decomposition. Thus, incorporating additional low-frequency components into the initial inference noise can significantly improve the quality of the generated video~\cite{si2024freeu,ren2024consisti2v,wu2023freeinit}. 

\begin{figure*}[ht]  
\centering
\subfloat[Input image]
  {
     \includegraphics[width=0.33\linewidth]{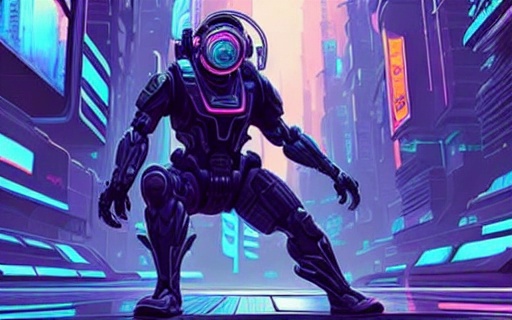}
  }
\subfloat[FFT]
  {
    \animategraphics[width=0.33\linewidth]{8}{videos/dct_vs_fft/fft/}{0}{15}
  }
\subfloat[DCT]
  {
    \animategraphics[width=0.33\linewidth]{8}{videos/dct_vs_fft/dct/}{0}{15}
  }

\caption{\textbf{Influence of the FFT and DCT decomposition.} The prompt is \textit{``a robot dancing''}. \textit{Best viewed with Acrobat
Reader. Please click the image to view the animated videos.}}
\label{fig_fft_vs_dct}
\end{figure*}

Previous approaches~\cite{ren2024consisti2v,wu2023freeinit} utilize Fast Fourier Transformation (FFT) to capture the low-frequency components and combine them with the initial inference noise during inference. However, as depicted in Fig.\ \ref{fig_fft_vs_dct}, directly applying the FFT-based decomposition method in our model is prone to causing color inconsistencies in the generated videos. Consequently, we introduce a frequency-domain decomposition strategy based on Discrete Cosine Transformation (DCT), referred to DCTInit. The principle behind it lies in the assumption of FFT, that the signal is periodic; if the real signal is not strictly periodic, truncation and periodic extension of the signal will introduce spectral leakage, resulting in inaccurate frequency-domain representations. In contrast, DCT assumes that the input signal is symmetrically extended, concentrating energy on low-frequency components, which is advantageous as coarse layout guidance in the inference stage of the image animation task. Mathematically, given the latent code $z_1$ of the input static image $x_1$ and the inference noise $\epsilon$, we first add $\tau$-step inference noise to $z_1$, leading to $z^{\tau}_1$. We then extract the low-frequency cosine transformation coefficients $\mathcal{D}_{z^{\tau}_1}^L$ of $z^{\tau}_1$ through the formulation $\mathcal{DCT}_{3D}(z^{\tau}_1) \odot \mathcal{H}$ and the high-frequency cosine transformation coefficients $\mathcal{D}_{\epsilon}^H$ of $\epsilon$ through the formulation $\mathcal{DCT}_{3D}(\epsilon) \odot (1-\mathcal{H})$, respectively. Finally, the refinement $\epsilon'$ is obtained through the formulation $\mathcal{IDCT}(\mathcal{D}_{z^{\tau}_1}^L + \mathcal{D}_{\epsilon}^H)$. Here, $\mathcal{DCT}$ represents the DCT operated on both spatial and temporal dimensions, $\mathcal{IDCT}$ represents the inverse DCT operation, and $\mathcal{H}$ represents the low pass filter. The refinement noise $\epsilon'$, which contains the low-frequency information of $z_1$, is then used for denoising. As shown in Fig.\ \ref{fig_effective_dctinit}, DCTInit can improve the temporal consistency and mitigate sudden motion changes in generated videos.


\section{Experiments}
\label{experiments}

\begin{figure*}[!t]
  \centering
  \begin{tabular}{c@{\hspace{0.1em}}c@{\hspace{0.1em}}c@{\hspace{0.1em}}c@{\hspace{0.1em}}c@{\hspace{0.1em}}c@{\hspace{0.1em}}c}
    
    \includegraphics[width=0.25\linewidth]{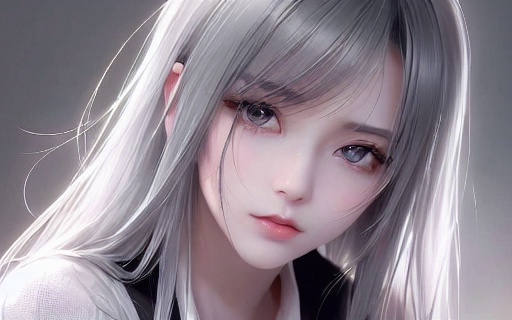} &
    \animategraphics[width=0.25\linewidth]{8}{videos/pikalab/}{0}{15} &
    \animategraphics[width=0.25\linewidth]{8}{videos/genmo/}{0}{15} &
    \animategraphics[width=0.25\linewidth]{8}{videos/consist2v/}{0}{15} \\
    Input Image & \href{https://www.pika.art/}{Pika Labs} & \href{https://www.genmo.ai/}{Genmo} & ConsistI2V~\cite{ren2024consisti2v} \\

    Prompt: \textit{``Girl smiling''} & \animategraphics[width=0.25\linewidth]{8}{videos/dynamicrafter/}{0}{14} &
    \animategraphics[width=0.25\linewidth]{8}{videos/i2vgen-xl/}{0}{15} &
    \animategraphics[width=0.25\linewidth]{8}{videos/seine/}{0}{15} \\
    & DynamiCrafter~\cite{xing2023dynamicrafter} & I2VGen-XL~\cite{zhang2023i2vgen} & SEINE~\cite{chen2023seine} \\

    & \animategraphics[width=0.25\linewidth]{8}{videos/pia/}{0}{15} &
    \animategraphics[width=0.25\linewidth]{8}{videos/svd/}{0}{13} &
    \animategraphics[width=0.25\linewidth]{8}{videos/ours/}{0}{15} \\
    & PIA~\cite{zhang2023pia} & SVD~\cite{blattmann2023stable} & Ours \\
  \end{tabular}
  \caption{\textbf{Qualitative visual comparisons between the baselines and our model}. We compare our approach with both closed-source commercial tools and research works. \textit{``Girl smiling''} means the used prompt when the method accepts it. \textit{Best viewed with Acrobat
Reader. Please click the image to view the animated videos.}}

  \label{fig:qualitative_comparison}
\end{figure*}

We first outline the experimental setup, covering datasets, baselines, evaluation metrics, and implementation details in Sec.\ \ref{experiments_experimental_setup}. Following that, we compare the experimental results with state-of-the-art in Sec.\ \ref{experiments_comparisons_with_state_of_the_art}. Finally, we present the analysis of our model in Sec.\ \ref{experiments_analysis}. More visual results can be seen on \href{https://maxin-cn.github.io/cinemo_project/}{project website}.

\subsection{Experimental setup.}
\label{experiments_experimental_setup}
\textbf{Datasets and implementation details.} We train our model on the WebVid-10M~\cite{bain2021frozen} and Vimeo25M datasets \cite{wang2023lavie}, which contains approximately 10 million and 25 million text-video pairs, respectively. For each video, we randomly sample 16 frames at a spatial resolution of $320 \times 512$ pixels, with a frame interval ranging from 3 to 10 frames. This sampling strategy provides a consistent motion intensity distribution as a conditional input to our model. We use the first frame as the input static image and train our model to denoise the motion residuals of the subsequent 15 frames. Following \cite{zhang2023adding}, we randomly drop input textual prompts with a probability of 0.5 to enable classifier-free guidance \cite{ho2022classifier} at the training stage. During inference, we use the DDIM sampler \cite{song2021denoising} with 50 steps and apply classifier-free guidance with a guidance scale of 7.5 to animate images in our experiments. The model architecture of Cinemo is identical to LaVie~\cite{wang2023lavie}, with the
addition of a linear layer for projecting motion intensity buckets $b$ into embeddings. The overall model is optimized using Adam on 8 NVIDIA A100 (80G) GPUs for one week, with a total batch size of 80. 

\textbf{Baselines and evaluation metrics.} We compare with recent state-of-the-art animation methods, including SVD \cite{blattmann2023stable}, I2VGen-XL \cite{zhang2023i2vgen}, DynamiCrafter \cite{xing2023dynamicrafter}, SEINE \cite{chen2023seine}, ConsistI2V \cite{ren2024consisti2v}, PIA \cite{zhang2023pia} and Follow-Your-Click \cite{ma2024follow}. Additionally, we compare our model against the commercial tools, \href{ https://runwayml.com/ai-magic-tools/gen-2/}{Gen-2}, \href{https://www.genmo.ai/}{Genmo}, and \href{https://www.pika.art/}{Pika Labs}. Following recent works \cite{ren2024consisti2v,dai2023animateanything}, we evaluate our model on two public datasets MSR-VTT \cite{xu2016msr} and UCF-101 \cite{soomro2012dataset}. We utilize Fr{\'e}chet Video Distance (FVD) \cite{unterthiner2019fvd} and Inception Score IS \cite{saito2017temporal} for assessing video quality, Fr{\'e}chet Inception Distance (FID) \cite{parmar2021buggy} for evaluating frame quality, and the CLIP similarity (CLIPSIM) \cite{wu2021godiva} for measuring video-text alignment. We evaluate FVD, FID, and IS on UCF-101 using 2,048 random videos, and FVD and CLIPSIM on the MSR-VTT test split, which consists of 2,990 videos. Our primary focus is to assess the animation capability of our model. We select a random frame from a video snippet sourced from our evaluation dataset. This frame, combined with a textual prompt, is used as the input to generate animated videos in a zero-shot manner.


\subsection{Comparisons with state-of-the-art} 
\label{experiments_comparisons_with_state_of_the_art}
\textbf{Qualitative results.} Animated video results can be found in Fig. \ref{fig:qualitative_comparison}. We find that our model creates animated videos that more accurately align with the prompt ``\textit{Girl smiling}''. While PIA also responds to the given prompt to some extent, its output significantly deviates from the input static image, failing to preserve the fine details of the original image. On the other hand, SEINE and SVD can create coherent and smooth videos but face challenges in matching the textual prompt. Meanwhile, videos generated by DynamiCrafter are nearly static, lacking dynamic changes. We observe that ConsistI2V and I2VGen-XL tend to deviate from the input static image and exhibit noticeable distortion in subsequent frames. While the commercial tool Pika Labs can maintain relatively high consistency with the input image, akin to the performance of SVD, Pika Labs is less sensitive to provided textual prompts. Another commercial solution, Genmo, tends to generate videos leaning towards a cartoon style, with little consistency with the original input image. All these observations collectively affirm that our model excels in producing coherent and consistent video content in response to specific textual prompts. 

\textbf{Quantitative results.} We conduct a comparative analysis on the evaluation metrics FVD, IS, FID, and CLIPSIM between our model and the baseline methods Gen-2, I2VGen-XL, DynamiCrafter, SEINE, ConsistI2V, and Follow-Your-Click on the UCF-101 and MSR-VTT datasets. All quantitative experiments are conducted in a zero-shot manner. From Tab. \ref{table_ucf_msrvtt}, we can see that our model gets the highest scores on the five metrics and achieves the best performance.

Based on the above analysis, our model demonstrates excellent performance in image animation quality and consistency. Compared to other baseline models, our model achieves significant improvements in all quantitative metrics.

\begin{table}[]
\caption{\textbf{Quantitative comparisons between the baselines and our model.} $\downarrow$ means the lower the better. $\uparrow$ means the higher the better.} 
\centering
\begin{tabular}{lccccc}
\toprule
\multirow{2}{*}{Method} & \multicolumn{3}{c}{UCF-101} & \multicolumn{2}{c}{MSR-VTT} \\ \cmidrule(lr){2-4} \cmidrule(lr){5-6} 
                        & FVD $\downarrow$     & IS $\uparrow$     & FID $\downarrow$    & FVD $\downarrow$         & CLIPSIM $\uparrow$     \\ \midrule
\href{ https://runwayml.com/ai-magic-tools/gen-2/}{Gen-2}                  & -        & -       & -      & 496.17       & -            \\
I2VGen-XL~\cite{zhang2023i2vgen}              & 597.42   & 18.20   & 42.39  & 270.78       & 0.2541       \\
DynamiCrafter~\cite{xing2023dynamicrafter}          & 404.50   & 41.97   & 32.35  & 219.31       & 0.2659       \\
SEINE~\cite{chen2023seine}                  & 306.49   & 54.02   & 26.00  & 152.63       & {0.2774}       \\
ConsistI2V~\cite{ren2024consisti2v}              & {177.66}   & {56.22}   & {15.74}  & {104.58}       & 0.2674       \\ 
Follow-Your-Click~\cite{ma2024follow}       & -        & -       & -      & 271.74       & -             \\  \midrule
Ours                    & \textbf{168.16}   &  \textbf{58.71}      & \textbf{13.17}       & \textbf{93.51}       & \textbf{0.2858}             \\    \bottomrule
\end{tabular}
\label{table_ucf_msrvtt}
\end{table}

\subsection{Analysis}
\label{experiments_analysis}
In this section, we conduct ablative studies and explore potential applications. Finally, we present limitations and discussions of our proposed model.

\begin{figure}[ht]  
\centering
\subfloat[Input image]
  {
     \includegraphics[width=0.25\linewidth]{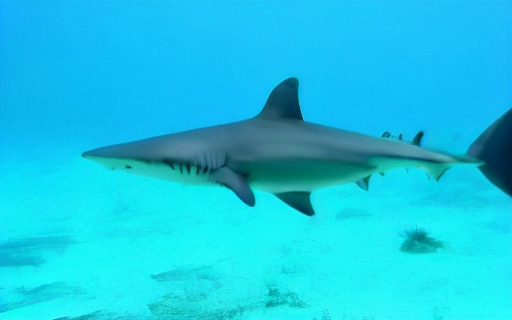}
  }
\subfloat[$b=0$]
  {
    \animategraphics[width=0.25\linewidth]{8}{videos/motion_bucket/0000/}{0}{15}
  }
\subfloat[$b=3$]
  {
    \animategraphics[width=0.25\linewidth]{8}{videos/motion_bucket/0003/}{0}{15}
  }
\subfloat[$b=6$]
  {
    \animategraphics[width=0.25\linewidth]{8}{videos/motion_bucket/0006/}{0}{15}
  } \\ 
\subfloat[$b=9$]
  {
    \animategraphics[width=0.25\linewidth]{8}{videos/motion_bucket/0009/}{0}{15}
  }
\subfloat[$b=12$]
  {
    \animategraphics[width=0.25\linewidth]{8}{videos/motion_bucket/0012/}{0}{15}
  }
\subfloat[$b=15$]
  {
    \animategraphics[width=0.25\linewidth]{8}{videos/motion_bucket/0015/}{0}{15}
  }
\subfloat[$b=18$]
  {
    \animategraphics[width=0.25\linewidth]{8}{videos/motion_bucket/0018/}{0}{15}
  }

\caption{\textbf{Motion intensity controllability.} The prompt is \textit{``shark swimming''}. Our model allows users to control the motion intensity by setting the input-associated information to different values. \textit{Best viewed with Acrobat
Reader. Please click the image to view the animated videos.}}
\label{fig_motion_intensity_control}
\end{figure}

\textbf{Motion intensity controllability.} Our model can control the motion intensity of the animated videos by setting the motion intensity bucket $b$ to different values. As shown in Fig. \ref{fig_motion_intensity_control}, our proposed SSIM-based strategy can effectively and smoothly adjust the intensity of motion. As the motion intensity bucket $b$ varies from 0 to 18, the motion intensity of the shark gradually increases (prompt: \textit{``shark swimming''}). This progression can be observed as the motion transitions of the shark from an initial slight tail wagging to a significant spatial displacement.

\begin{figure*}[ht]  
\centering
\subfloat[Input image]
  {
     \includegraphics[width=0.33\linewidth]{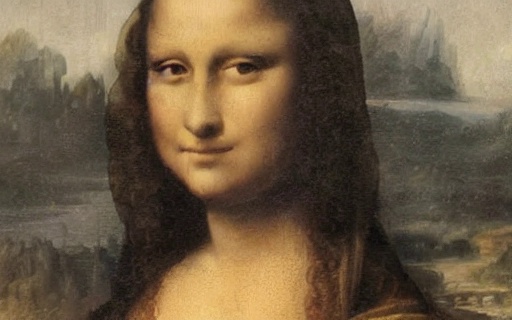}
  }
\subfloat[W/O DCTInit]
  {
    \animategraphics[width=0.33\linewidth]{8}{videos/effect_dctinit/wo_dctinit/}{0}{15}
  }
\subfloat[Ours]
  {
    \animategraphics[width=0.33\linewidth]{8}{videos/effect_dctinit/w_dctinit/}{0}{15}
  }

\caption{\textbf{Effectiveness of DCTInit.} The middle video is generated by our model without enabling DCTInit. The prompt is \textit{``woman smiling''}. \textit{Best viewed with Acrobat
Reader. Please click the image to view the animated videos.}}
\label{fig_effective_dctinit}
\end{figure*}

\textbf{Effectiveness of DCTInit.} Our experimental results demonstrate that the proposed DCTInit can stabilize the video generation process and effectively mitigate sudden motion changes. As shown in Fig. \ref{fig_effective_dctinit} (b), our model can still create reasonable videos that fit the expected scenarios without enabling DCTInit, but occasionally, there are instances of abrupt motion changes. Therefore, the role of DCTInit is to help the model generate videos with smooth and natural motion transitions. Additionally, in Fig. \ref{fig_fft_vs_dct}, we also demonstrate that the DCT frequency domain decomposition strategy can effectively mitigate the color inconsistency issues caused by the FFT frequency domain decomposition method detailed as in Sec. \ref{method_dct_based_noise_refinement}.

\textbf{Motion control by prompt.} We aim to learn the distribution of motion residuals rather than directly predicting subsequent frames. As depicted in Algorithm. \ref{alg:training}, information from the input static image is integrated into the model in a novel manner. These designs strengthen the connection between the predicted motion residuals and the input static image, as well as significantly improve the alignment accuracy between the video content and the given text descriptions.

\begin{figure*}[ht]  
\centering
\subfloat[Input image]
  {
     \includegraphics[width=0.25\linewidth]{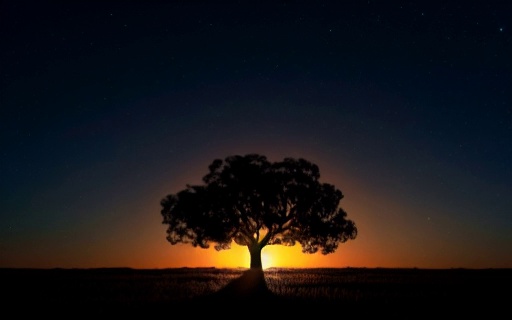}
  }
\subfloat[\textit{``fireworks''}]
  {
    \animategraphics[width=0.25\linewidth]{8}{videos/motion_control/fireworks/}{0}{15}
  }
\subfloat[\textit{``leaves swaying''}]
  {
    \animategraphics[width=0.25\linewidth]{8}{videos/motion_control/leaves_swaying/}{0}{15}
  }
\subfloat[\textit{``lightning''}]
  {
    \animategraphics[width=0.25\linewidth]{8}{videos/motion_control/lightning/}{0}{15}
  }

\caption{\textbf{Motion control by textual prompts.} (b), (c), and (d) use \textit{``fireworks''}, \textit{``leaves swaying''}, and \textit{``lightning''} as textual prompts, respectively. Our model can effectively respond to textual prompts, leading to visually appealing outcomes. \textit{Best viewed with Acrobat
Reader. Please click the image to view the animated videos.}}
\label{fig_motion_vary_control}
\end{figure*}

As shown in Fig. \ref{fig_effective_dctinit}, \ref{fig_motion_intensity_control} and \ref{fig_motion_vary_control}, our model does not rely on complex guiding instructions, which are preferred by users. Experiments have demonstrated that even simple textual prompts can yield satisfactory visual effects. Furthermore, as illustrated in Fig. \ref{fig_motion_vary_control}, our model can flexibly respond to textual prompts by incorporating new elements into the generated video, leading to outcomes that are both compliant with specifications and visually appealing.

\begin{figure*}[ht]
\centering
\begin{tabular}
{c@{\hspace{0.1em}}c@{\hspace{0.1em}}c@{\hspace{0.1em}}c@{\hspace{0.1em}}c@{\hspace{0.1em}}c@{\hspace{0.1em}}c}
Original video & First frame & Edited first frame & Output video \\
\animategraphics[width=0.25\linewidth]{8}{videos/video_editing/a_man_walking_on_the_beach/}{0}{15} &
\includegraphics[width=0.25\linewidth]{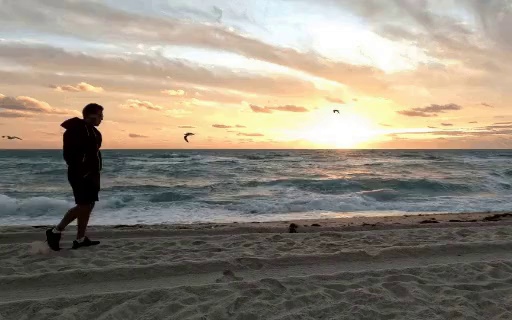} &
\includegraphics[width=0.25\linewidth]{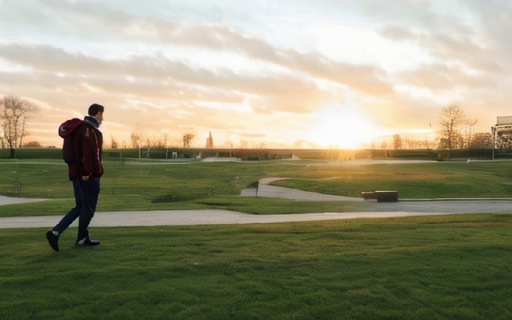} &
\animategraphics[width=0.25\linewidth]{8}{videos/video_editing/editing_a_man_walking_in_the_park/}{0}{15} \\
\animategraphics[width=0.25\linewidth]{8}{videos/video_editing/a_corgi_walking_in_the_park_at_sunrise_oil_painting_style/}{0}{15} &
\includegraphics[width=0.25\linewidth]{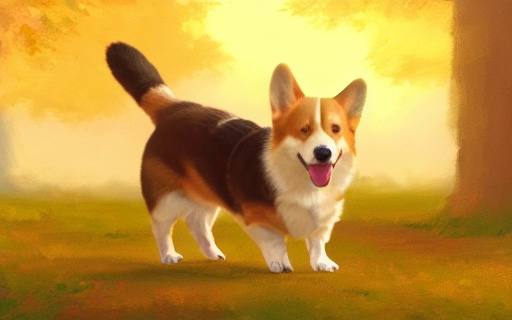} &
\includegraphics[width=0.25\linewidth]{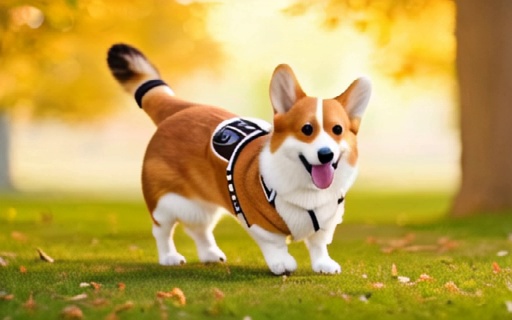} &
\animategraphics[width=0.25\linewidth]{8}{videos/video_editing/editing_a_corgi_walking_in_the_park_at_sunrise,_oil_painting_style/}{0}{15} \\
\end{tabular}

\caption{\textbf{Motion transfer/Video editing results.} Our model can easily extend to other applications. \textit{Best viewed with Acrobat
Reader. Please click the image to view the animated videos.}}
\label{fig_motion_transfer_video_editing}
\end{figure*}

\textbf{Motion transfer/Video editing.} The uniqueness of our approach lies in its focus on learning the distribution characteristics of motion residuals, rather than directly predicting the next frame. This advantage allows us to utilize the DDIM inversion algorithm~\cite{wallace2023edict} to get the initial inference noise corresponding to the motion residuals of the given video. Subsequently, we edit the first frame using the off-the-shelf image editing technique~\cite{tumanyan2023plug}. By inputting the edited first frame and the corresponding initial inference noise of the motion residuals into our model, we can achieve other applications such as motion transfer and video editing as shown in Fig. \ref{fig_motion_transfer_video_editing}.

\textbf{Limitations and discussions.} Our model is based on the pre-trained LaVie~\cite{wang2023lavie} model and is further trained on similar datasets. This means that the performance of our model is, to some extent, limited by the inherent characteristics of LaVie. For example, the resolution of the current video generation is constrained by LaVie, fixed at 320 x 512. In recent years, the technological development trend in the field of video generation has clearly shifted towards Transformer-based architectures, gradually replacing the traditional UNet architecture. This shift is mainly due to the more effective scalability of the model parameters of Transformers. In light of this, our future plans include adopting Transformer-based architectures, such as Latte~\cite{ma2024latte}, to further validate and optimize our model.

\section{Conclusion}
\label{conclusion}

In this paper, we introduce Cinemo, a novel image animation model that achieves both image consistency and motion controllability. To accomplish this, we focus on learning the distribution of motion residuals rather than directly predicting the next frames. Additionally, we implement an effective SSIM-based strategy to control motion intensity in the generated videos. During inference, we propose DCTInit, which utilizes the low-frequency Discrete Cosine Transform coefficients of the input static image to refine initial noise and stabilize the generation process. Our experimental results demonstrate the effectiveness and superiority of our approach compared to existing baselines.

\bibliographystyle{unsrt}
\bibliography{bibliography}







\end{document}